%% file: main.tex
\documentclass{article}

\usepackage{microtype}
\usepackage{graphicx}
\usepackage{subcaption}
\usepackage{booktabs} 

\usepackage{hyperref}

\usepackage[accepted]{icml2026}

\usepackage{amsmath}
\usepackage{amssymb}
\usepackage{mathtools}
\usepackage{amsthm}
\usepackage{xcolor}

\usepackage[capitalize,noabbrev]{cleveref}

\theoremstyle{plain}

\theoremstyle{definition}

\theoremstyle{remark}

\usepackage[textsize=tiny]{todonotes}

\icmltitlerunning{Position: the Stochastic Parrot in the Coal Mine}

\begin{document}

\twocolumn[
  \icmltitle{Position: the Stochastic Parrot in the Coal Mine\\Model Collapse is a Threat to Low-Resource Communities}



  \icmlsetsymbol{equal}{*}

  \begin{icmlauthorlist}
    \icmlauthor{Devon Jarvis}{equal,wits,mind}
    \icmlauthor{Richard Klein}{wits,mind}
    \icmlauthor{Benjamin Rosman}{mind,wits}
    \icmlauthor{Steven James}{wits,mind}
    \icmlauthor{Stefano Sarao Mannelli}{chalmer,wits}
  \end{icmlauthorlist}

  \icmlaffiliation{wits}{School of Computer Science and Applied Mathematics, University of the Witwatersrand, Johannesburg, South Africa}
  \icmlaffiliation{mind}{Machine Intelligence and Neural Discovery Institute, University of the Witwatersrand, Johannesburg, South Africa}
  \icmlaffiliation{chalmer}{Data Science and AI, Computer Science and Engineering, Chalmers University of Technology and University of Gothenburg}

  \icmlcorrespondingauthor{Devon Jarvis}{devon.jarvis@wits.ac.za}

  \icmlkeywords{Machine Learning, ICML}

  \vskip 0.3in
]



\printAffiliationsAndNotice{}  

\begin{abstract}
\looseness=-1
Model collapse, the degradation in performance that arises when generative models are trained on the outputs of prior models, is an increasing concern as artificially generated content proliferates. Related critiques of large language models have highlighted their tendency to reproduce frequent patterns in training data, their reliance on vast datasets, and their substantial environmental cost. Together, these factors contribute to data degradation, the reinforcement of cultural biases, and inefficient resource use. In this position paper we aim to combine these views and argue that model collapse threatens current efforts to democratize AI. By reducing training efficiency and skewing data distributions away from the tails of their support, model collapse disproportionately impacts low-resource and marginalized communities. We examine both the environmental and cultural implications of this phenomenon, situate our position within recent position papers on model collapse, and conclude with a call to action. Finally, we outline initial directions for mitigating these effects.
\end{abstract}

\section{Introduction} \label{Sec:Introduction}
The emergence of generative artificial intelligence (AI) techniques, such as transformers \citep{vaswani2017attention} and diffusion models \citep{rombach2022high}, has led to widespread public adoption in recent years.
ChatGPT alone is estimated to generate approximately $0.1\%$ of all words produced globally each day \citep{schaeffer2025position}.
This adoption is already reshaping how people work and learn \citep{chen2020artificial,brynjolfsson2025generative}. While generative AI has the potential to be broadly democratising, the consequences of its large-scale use remain poorly understood \citep{kosmyna2025your,stankovic2025comment}.

\looseness=-1
The low cost of content generation has led to an unprecedented growth in available data \citep{wei2024understanding}, with recent estimates suggesting that generated content constitutes roughly 30\% of the internet as of 2025 \citep{spennemann2025delving}. As a result, successive generations of generative models are trained on increasing amounts of AI-generated data. Several recent studies show that training repeatedly on such ``synthetic'' data degrades generation quality across iterations \citep{zhu2024synthesize,shumailov2024ai,dohmatob2024model,seddik2024bad}. This phenomenon, known as model collapse, leads to highly repetitive or incoherent text and overly stylized, unrecognisable images \citep{bohacek2023nepotistically,dohmatob2024tale,gerstgrasser2024model}.

Parallel work has also raised concerns around the propensity of language models to reinforce stereotypes and biases present in their training data, and the significant computational resources required to train them \citep{bender2021dangers,samsi2023words,jegham2025hungry}. Due to the expense in collecting data from communities less prevalent on the internet \citep{kandpalposition} and training such extremely large models, low-resource communities have less representation in emerging technologies which hinders the applicability of the technology to their contexts. 
 
At the same time, the energy demands of large-scale computation continue to drive new infrastructure development in already energy-intensive regions \citep{electricity2024analysis}, while the environmental consequences are borne globally. Low-resource communities are particularly vulnerable to these effects, including increased exposure to climate-related disasters \citep{van2006impacts}. As a result, these communities bear a disproportionate share of the costs while receiving comparatively little of the benefit \citep{farnadi2024position}. As \citet{strubell2019energy} note, marginalized communities that experience the effects of environmental change first are often the last to gain access to new technologies.\\

\looseness=-1
In the context of language models, these structural imbalances are compounded by the phenomenon described by \citet{bender2021dangers} as ``Stochastic Parrots''. While large language models can produce fluent and coherent text when trained on vast datasets, their outputs remain ungrounded, lacking social context and adherence to human norms. The consequence is that the models repeat statements based on frequency rather than merit or accuracy \citep{orgad2025llms}. Thus, generated text can be skewed by social constructs that are potentially harmful and factually incorrect. Even in the case of hallucinations, the errors are not arbitrary and reflect an underlying bias of the models against under-represented groups \citep{farnadi2024position, orgad2025llms}. Coupled with the human proficiency at finding meaning as an interlocutor \citep{clark2004changing,clark2004speaking,janzen2008intersubjectivity}, this creates systems that can be misleading or harmful at scale \citep{choi2024llm}. Although we revisit the continued relevance of the Stochastic Parrot analogy after several years of progress in Sec.\ref{Sec:Alternative}, it remains foundational to the concerns raised in this perspective.

\looseness=-1
While model collapse, the energy consumption of Large Language Models (LLMs) and their propensity to repeat hegemonic or biased viewpoints are pressing issues in isolation, their interaction has received little attention. We begin to close this gap by arguing that \textbf{model collapse amplifies the existing harms of LLMs for low-resource languages and marginalized communities by producing models which favour hegemonic viewpoints more with each generation, remove under-represented viewpoints from the data distribution and scale poorly with the addition of real data. By poisoning the data and removing the tails of the distribution, it undermines ongoing efforts to democratize AI. Addressing model collapse should therefore be a central concern for research on low-resource settings and AI fairness.}

\looseness=-1
To substantiate this claim, we begin in Sec.\ref{Sec:Background} by reviewing the recent theoretical and empirical findings on model collapse. In Sec.\ref{Sec:Parrots} we reconsider the points raised in the seminal work by \citet{bender2021dangers} on Stochastic Parrots in light of the new findings on model collapse \citep{dohmatob2024tale,shumailov2024ai}. We highlight how model collapse exacerbates both the environmental and social risks associated with large language models, while also introducing new risks. Sec.\ref{Sec:Alternative} provides some alternative views which might mitigate the potential risks of model collapse on low resource settings. Sec.\ref{Sec:Call} concludes with a call to action and initial recommendations for immediate directions of research  to  better understand or mitigate the risks of model collapse for low-resource and marginalized communities. 

Throughout this work we make use of the term ``low-resource community'' and it is important to specify what we mean. In this case, the term is not limited to reflecting the fact that the community has a relatively smaller proportion of their data present in datasets or on the internet, although this is part of it. Through this term we also include communities which lack sufficient compute or financial means to train large models or collect data freely, relative to the dominant communities on the internet. Furthermore, this notion of a dominant community is highly context specific and temporal. A community may be dominant is a particular dataset and not within another. Importantly, there are a number of reasons why a community's data may occupy the tails of a data distribution and we do not consider any one reason here. We do, however point to a few beyond a lack of representation throughout this work. Our point here is not to group all of these concerns into one, but to argue that these reasons and difficulties exist disproportionately for some communities which may lead to more severe model collapse for these same communities. Thus, the term ``low-resource community'' is used for ease of communication and not intended to convey a uniform group. Understanding how model collapse manifests for different reasons is something we touch on in our call-to-action in Sec.\ref{Sec:Call}. Additionally, Tab.\ref{tab:summary-table} summarizes the main mechanisms of model collapse discussed throughout the paper, linking each cause to its expected effect on low-resource communities.

\textbf{Conflict of Interest Disclosure}: The authors have no conflict of interest to disclose.


\section{Background} \label{Sec:Background}
\input{summary-table}
\looseness=-1
While the precise definition of model collapse is still evolving and often conflates distinct effects \citep{schaeffer2025position}, the literature generally groups several phenomena under this heading, ranging from a catastrophic divergence in test loss -- often an artefact of unrealistic ``replace'' training paradigms \citep{schaeffer2025position} -- to subtler deformations of the data distribution. Despite this definitional ambiguity, the core concern generally refers to a specific failure mode: \textit{a progressive decrease in the marginal value of additional training data (when a proportion of that data is synthetic) and a simultaneous reduction in the diversity and quality of the output distribution} \citep{shumailov2024ai}.

This degradation is best understood through the lens of neural scaling laws that describe the predictable power-law improvement in model performance (usually measured by cross-entropy loss) as compute, dataset size, and parameter count increase \citep{kaplan2020scaling,bahri2024explaining}. These laws provide an optimistic alternative to the performance plateaus typical of overfitting in earlier architectures \citep{cogswell2015reducing,santos2022avoiding}. However, the introduction of synthetic data fundamentally alters this trajectory. Theoretical and empirical works indicate that at sufficient scales, even a small fraction of synthetic data (around 1\%) causes models to suffer from a change in scaling exponents -- essentially hitting a ``glass ceiling'' where larger datasets no longer yield proportional improvements \citep{dohmatob2024strong,dohmatob2024model,cui2025solvable}. This phenomenon is explained as the model overfitting to the dominant factors of variation in the original dataset which are then over-represented in the generated samples \citep{shumailov2024ai}. The generated samples enter the training data for subsequent models, strengthening these dominant factors further and creating a feedback loop where the model progressively loses the ability to represent the variability and structure of real-world data \citep{shumailov2023curse,bohacek2023nepotistically}.

This over-representation of dominant factors leads directly to a loss of variability through this feedback loop that systematically erases rare or ``tail'' examples \citep{shumailov2023curse,bohacek2023nepotistically}. Consequently, the rich variance of the original real-world distribution is replaced by a smoothed, homogenized approximation. The threat of model collapse is, therefore, not necessarily that models will become gibberish, but that they will become narrow—capable of reproducing frequent patterns but increasingly blind to the long tail of human diversity \citep{schaeffer2025position,shumailov2024ai}.\\

This process is closely related to a similar concept in cognitive science known as iterated learning (IL) which has been used to explain the refinement of language and behaviours over generations of human learners \citep{smith2003iterated}. In humans, the result is a highly structured (termed regular) language or set of actions that are easy to learn and can be composed to generalize systematically \citep{kirby2008cumulative}. However, in both empirical experiments and theoretical models, expressivity can be lost without external intervention and the differences between closely related data points is forgotten \citep{kirby2008cumulative,kirby2014iterated,jarvis2026compositionality}. Thus, to consider the inductive bias of the model to focus on dominant factors of variation as ``only negative'' is not grounded in the broader literature on cognitive modelling and machine learning. It is plausible that the real-world situation is more optimistic than those shown on smaller (but still naturalistic) datasets and in theoretical models. This is a point we return to in Sec.\ref{Sec:Alternative}.  However, when coupled with the existing biases of LLMs, such as promoting the hegemonic viewpoints in its training dataset \citep{bender2021dangers}, and the already enormous costs involved in training them \citep{xia2024understanding,kandpalposition}, it is important to consider the potential harms which model collapse may cause for low-resource and marginalized communities.



\section{Generations of Stochastic Parrots} \label{Sec:Parrots}


\subsection{Environmental and Financial Costs} \label{Sec:Environment}

The financial costs associated with training LLMs are often scrutinized \citep{strubell2019energy}; for example, training a single BERT base model without hyperparameter tuning on GPUs was estimated to consume as much energy as a flight across America \citep{strubell2019energy}, training a large transformer model with neural architecture search emits as much $CO_2$ emissions as the average human over 60 years \citep{vaswani2017attention,bender2021dangers}, and training GPT-4 reportedly cost over \$100m \citep{odonnell2025math}. 
However, this estimate does not consider the cost of the data used to train the models (often because it goes unpaid). Based on conservative estimated of wage rates, the costs of obtaining the training datasets are 10--1000 times larger than the costs to train the models themselves \citep{kandpalposition}. Even if the real cost is highly discounted, there is still an opportunity cost and loss of human capital associated with the use of such expensive data.

\looseness=-1
In many cases, the high cost of training an LLM is justified by its use as a foundation model \citep{xu2024survey}. Hence, the cost is amortized over its entire life-cycle and uses \citep{ren2024reconciling}. However, most research establishing the costs of training LLMs is based (necessarily) in idealized settings and with clean datasets \citep{zeighami2025cut}. For example, \citet{strubell2019energy} estimate that an increase in BLEU score of 0.1 from using neural architecture search for English to German translation results in an increase of \$150,000 in compute costs. Further, this cost profile  shifts drastically with new modalities: generating a five-second video can consume more than 700 times the energy of generating an image \citep{odonnell2025math}. Consequently, the compute cost of training the largest deep learning models increased roughly 300,000-fold between 2015 and 2021 \citep{devlin2019bert}, a trend expected to accelerate with large-scale infrastructure projects such as the \$500 billion ``Stargate'' initiative \citep{odonnell2025math}. 

Using the linear model of model collapse proposed by \citet{dohmatob2024strong}, achieving a tenfold reduction in test loss (from $10^{-1}$ to $10^{-2}$) requires an order-of-magnitude increase in dataset size (from $10^3$ to $10^4$), consistent with established power-law scaling \citep{kaplan2020scaling}.
Considering a dataset where $10\%$ is synthetic, the increase in dataset size needed is \emph{another} order of magnitude (from $10^3$ to $10^5$). While it is important to recognize that this is a theoretical model (including a hyperparameter to control the quality of the synthetic data), even if the model is pessimistic, model collapse can only stand to increase the amount of data and compute needed to train new models. Coupled with the point that the hidden costs of producing the training data are often overlooked \citep{kandpalposition}, this increase in cost becomes even more severe. 

While the natural answer of collecting more real data may be viable in some settings, if the cost of obtaining data is even on the lower end of the estimate by \citet{kandpalposition}, then this will likely not be a viable solution for many populations. This leaves low-resource communities unable to obtain data without significant cost or without the use of synthetic data as a mechanism to supplement datasets. Thus, when benchmarking the environmental cost of training foundation models it is important to also consider the effect of cheap synthetic data or expense of real data on this process, or else it stands to be highly optimistic. This inequity is compounded by the fact that data centres are often clustered in regions with carbon-intensive grids, resulting in emissions 48\% higher than the national average \citep{odonnell2025math}.

While it is unrealistic to expect uniform technological advancement across different communities and sub-fields, the typical belief is that progress in one space will lend insight and lead to advancement in another. Instead, model collapse may result in fundamentally uneven access to new technology between communities. Moreover, the communities most hindered are precisely those that could benefit the most from technological advancement. Even under more optimistic scenarios, model collapse threatens to undermine ongoing efforts to reduce the environmental impact of training LLMs \citep{huangtowards,iftikhar2025enhancing,shi2025efficient} and the injustice this causes \citep{strubell2019energy}. Thus, beyond unequal access to new technology and the disproportionate impact of environmental harm on certain sub-populations, we hypothesize that model collapse and the cost of data acquisition could introduce a third disadvantage to low-resource communities: \textit{a lower upper-bound on technological advancement} from these models.

\subsection{Cultural Costs} \label{Sec:Culture}

The source of injustice from LLMs is not only in the resources required to train and use the models, however. There are also valid concerns stemming from the effect of LLMs on public perception and culture. Foundational to this section is the recognition that model collapse results in a lack of diversity of generated content and that what remains in the data is the highly probable content from the original dataset \citep{alemohammad2023self,bertrandstability}. Thus, data which would occupy the tails of a probability distribution are forgotten or corrupted \citep{shumailov2024ai}. Datasets used to train generative models are typically scraped from the internet, which has been shown to be problematic \citep{basta2019evaluating,birhane2021large,de2019does,kurita2019measuring}. This has resulted in the models encoding biases and derogatory associations in some cases \citep{bao2024decoding,shin2024ask}. Even when there is not overt harm, the models favour the hegemonic viewpoint presented on the internet sites where the data is scraped from \citep{bender2021dangers} -- one of the foundational points leading to the stochastic parrot phrasing.

The comprehension of LLMs regarding minority groups has often been shown to be limited, absent, or stereotyped. \citet{farnadi2024position} comprehensively analyse the types of discrimination LLMs perform, identifying not only privacy issues --- as LLMs tend to memorize under-represented data more --- but also ``hallucination gaps'' where models hallucinate more facts regarding minority groups. Furthermore, they note ``underspecification disparities'' where models generate seemingly arbitrary text when addressing topics related to marginalized groups \citep{farnadi2024position}. Consequently, LLMs flatten complex concepts \citep{gallegos2024bias}, often aligning towards Western and English-speaking values \citep{rao2023ethical}. Together, these factors produce an outcome that amplifies bias and removes diversity. This effect is further amplified by standard compression and acceleration techniques --- including knowledge distillation, quantization, pruning, and caching --- which have been shown to further increase bias \citep{silva2021towards, ahn2022knowledge, kirsten2025impact}. Cases of biased data generation are also not limited to LLMs. For example, it has been shown that stable diffusion has a propensity to perpetuate stereotypes regarding race and skin tone which has a homogenising effect on less frequently represented racial and ethnic groups \citep{wilson2025bias}.

Data curation has been proposed as a potential solution to this problem \citep{birhane2021large,jo2020lessons}, but there are a number of ways that model collapse can undermine the curation process. If mitigating model collapse becomes a part of data curation, then it will be necessary to detect synthetic data for its removal. However, this is becoming an extremely difficult task as generative models become more convincing to human perception \citep{boutadjine2025human}. Thus, curation will rely more on automated detection systems. Although there has been some success in automating data mixing and curation through methods like instruction-tuning \citep{jiang2024instruction,raj2024breaking,liu2025regmix,thudimixmin} these approaches are still potentially fallible, uninterpretable and require large quantities of real or synthetic data to train \citep{chaka2023detecting,elkhatat2023evaluating,ghiuruau2024distinguishing}. This could stand to worsen the  of curation to remove data from marginalized groups or those who express fair but less frequent views. For example, the Colossal Clean Crawled Corpus \citep{raffel2020exploring} was cleaned by discarding any sample with one of 400 bad words. While this is effective at removing some of the worst known biases and slurs from data, it also reverses any attempts to reclaim works by marginalised groups \citep{bender2021dangers}. This effect of LLMs to re-establish outdated connotations for words is termed ``value-lock'' \citep{bender2021dangers} and disturbs the role of language in shaping social norms and culture. As \citet{twyman2017black} note, a central aspect of recent social movements revolves around using online forums to document events, promote coverage of new events and marginalized perspectives, and change the perspectives on existing knowledge in light of new information. 

Value-lock was already a concern for first-generation LLMs based only on generated content being more likely to represent previously dominant (and hence frequently represented in the training data) viewpoints. The likely outcome of model collapse is that these dominant viewpoints will become even more over-represented in the dataset as subsequent generations of models are trained \citep{whitney2024real}. Equally, the less frequently encountered language which is deliberately used to change social perception will be forgotten over the generations. \citet{wyllie2024fairness} contextualize this within the rapid increase of LLM-generated text, defining the \textit{``Fairness Feedback Loop.''} They argue that model collapse is not merely a quality issue; it is an erasure engine that systematically removes under-represented communities from the digital reality defined by AI models. All of this stands to reinforce existing regimes of power and undermine attempts to promote social change. 

Importantly, the looming harm from model collapse is not limited to a shift of the data distributions towards previous views, as shown by recent cases in high-resource settings. For example, \citet{wang2025bias} show that GPT-2 favours right-leaning media when successively fine-tuned on a balanced media dataset from the US such that it undergoes model collapse. \textit{Thus, even in ideal settings model collapse has the propensity to inject bias where there seemingly is none, or based on underlying biases which are not immediately obvious to the expert and public raters of their dataset.}

\subsection{Transfer Learning} \label{Transfer}

Recent work on low-resource languages has increasingly relied on transfer learning from multilingual models \citep{lai2023chatgpt,devlin2019bert}. The rationale is that combining data from multiple languages—particularly high-resource ones—provides sufficient data to learn a semantically rich embedding space. The low-resource language can then leverage this space, enabling the model to represent meaning more effectively despite limited direct data. However, fine-tuning in this manner introduces several issues, including language drift, where the embedding space is progressively forgotten as the model is trained on a more narrowly defined task \citep{lee2019countering,lu2020countering}. In other words, while transfer learning provides benefits, the model can lose capabilities originally acquired during pretraining on a rich embedding space.

\setlength{\belowcaptionskip}{-20pt}
\begin{figure}[h!]
    \centering
    \includegraphics[width=0.85\linewidth]{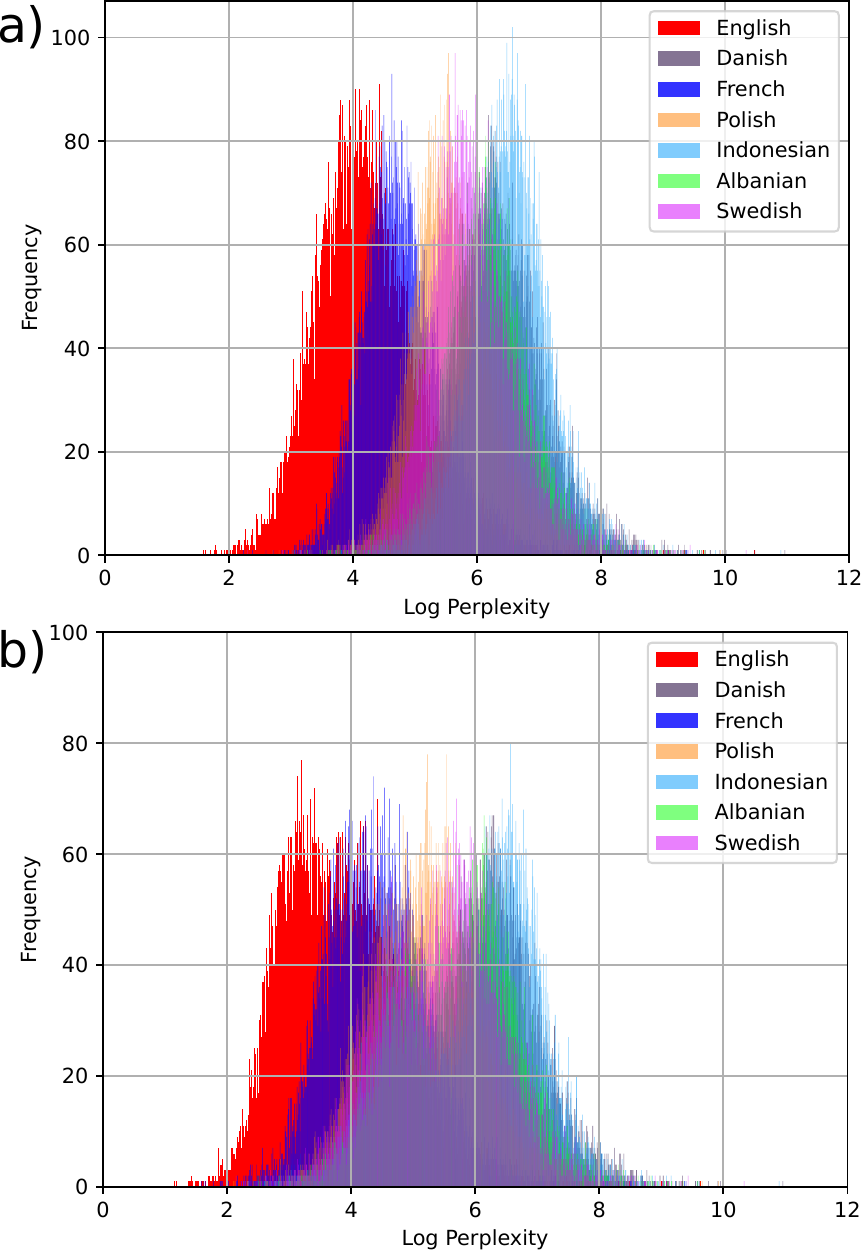}
    \caption{Perplexity of multiple languages using the Latin alphabet (potentially with some added characters) calculated using a pretrained GPT-2 \citep{radford2019language}. We present this result to support our position only. \\ a) \textbf{Original Distribution:} Note how the lower-resource languages occupy a distribution closer to the tails (at a higher perplexity) than the more high-resource languages such as English. Each language distribution is calculated using 15000 input sentences \citep{agentlans} and show log-perplexity for visibility.\\ b) \textbf{Collapsing Distribution:} Note how the low-resource languages shift the most when 5000 samples are generated for each language and added to the distribution with replacement. Swedish and Polish are affected worse than the other languages, while all distributions shift left and begin to overlap -- highlighting the homogenising effect of collapse.}
    \label{fig:perplexity}
\end{figure}
\setlength{\belowcaptionskip}{0pt}

How model collapse manifests in this setting where only data at the edge of the distribution (the low-resource language) is being generated is an open question. If the model remains multi-lingual, then model collapse will likely result in the erosion of low-resource languages from the data over time, and the model will progressively deteriorate faster on these languages. Figure~\ref{fig:perplexity} demonstrates this by showing the distribution of log-perplexity values GPT-2 \citep{radford2019language} attributes to 16000 sentences from various languages using the Latin alphabet \citep{agentlans}. Clearly, high-resource languages such as English and French have a lower initial perplexity compared to Indonesian and Albanian. By our current understanding of model collapse, these languages at the tails of the combined distribution should be forgotten first, which is supported by the initial shift in distributions when just 2000 generated samples are added. While this is a preliminary, it is still a stark result demonstrating more need for work focusing on the effect of model collapse on low-resource languages. 

Multi-lingual models are also often told which language to generate text for (or at least prompted in a particular language) and so it is unlikely to collapse in the same manner as if it was generating text from its intrinsic distribution of language. This deliberate mechanism introduced by the human user is a means by which model collapse could be mitigated. However, prompting and guiding the generation in this manner reintroduces a mechanism for bias \citep{yang2023adept,hida2024social,chisca2024prompting}. These and many more questions, such as the effect of model collapse on languages where code-switching is common \citep{woolford1983bilingual} (and itself a lingering issue for many low-resource languages \citep{yoo2025code,yoo2025code2,mohamed2025lost}) remain open and are important to fully grasp the effect model collapse could have on marginalized and low-resource communities.

This leads to our final consideration: the deliberate use of generative models to produce content for low-resource settings and marginalized communities. One reason this is needed is to ensure that this data remains represented in the data distributions of multi-lingual models to some degree. In essence, having synthetic data for your community enter into the broader distribution is likely better than leaving your community's data to slide further into the tails. Implicitly, there is a race between communities to keep pace with the rate of data generation of the others. Real data is certainly the best content to produce, but this favours communities with resources, as noted in Sec.\ref{Sec:Environment}. The net result is that marginalized communities may be pushed to poison their data merely to ensure their data remains salient in combined datasets. The other reason marginalized communities may still need to use generative models, and likely the larger reason, is to keep pace with the growing productivity internationally which can result from new technologies. Thus, these communities will need to adopt generative models to remain economically active within a globalized society with blurred cultural boundaries.

\looseness=-1
The big question is then: how quickly will content be generated in these settings with AI, compared to human content creation? In the best case, content generation will follow the same distribution of general content on the internet. In this case, all communities will have roughly the same percentage of synthetic content in their training data at a given time and will face a similar degradation from model collapse. Since the plateau from model collapse gets worse as more data is added \citep{shumailov2024ai}, this may even make model collapse initially worse for high-resource settings. However, it seems plausible that the technology will provide a proportionally larger effect on low-resource settings as the cost of generating content becomes significantly cheaper. Essentially, synthetic content generation is democratised far quicker than the internet as a whole. In this case the rate of generated content production will more closely follow the population of a country, while the rate of real content generation will align to resource access. Thus, in absolute terms, the amount of content generated by low-resource communities will be less but the proportion of synthetic data in these languages will be greater. \textit{This will result in low-resource communities encountering severe model collapse far quicker than high-resource communities and the erosion of these cultures, before we have the methods to reverse the collapse}.


\section{Alternative Views} \label{Sec:Alternative}
\looseness=-1
There are a number of alternative views to be considered, ranging from the problem being real but mitigated, to model collapse (or at least the bottleneck imposed by generations of trainers) being beneficial. Considering the former, the extent of model collapse is still being debated and some work does show that models can stabilize after a certain number of generations \citep{kazdan2024collapse}. Similarly, cases where extreme collapse are shown tend to be on rather small datasets and measured using metrics like the perplexity of individual samples \citep{shumailov2023curse}. While these metrics and experiments match qualitative evaluations of the generated content, how to accurately measure or characterize model collapse remains a valuable question. Due to this difficulty, isolating its effects is also difficult. A subtle but important alternative view may consider whether model collapse is really accelerating or introducing bias in the data distributions or just a symptom of a growing gap between communities based on access to resources. Thus, it is plausible that with sufficient real data, models will not be overwhelmed by synthetic data and that the data distributions may stabilize to content which is no more biased than it otherwise would be in the first generation \citep{alemohammad2023self,kazdan2024collapse}.

\looseness=-1
The comparison of model collapse to iterated learning supports the latter view that model collapse may even be beneficial. As noted in Sec.\ref{Sec:Background}, the generational transmission of language and the regularity this affords is a benefit for human language use and cognition \citep{kirby2008cumulative}. One could argue that model collapse will result in the loss of invalid information (which includes biases) due to the fact that it is less grounded within the environment.\footnote{We certainly run the risk of oversimplification here, as the degree to which biases are ingrained into the environment once they exist is a subtle point beyond the scope of this discussion.} As a result, the language which will be most represented will be grounded language which expresses truths about nature. In a sense, there are many ways to be wrong and so no one way should dominate the data distribution. 

It is important to note that the iterate learning algorithm can also result in a loss of expressivity when run with humans and some artificial agents \citep{kirby2008cumulative,kirby2014iterated,jarvis2026compositionality}. What is required is a mechanism for the reintroduction of words and forgotten information and some creativity from the speakers. It is conceivable that with such a mechanism, LLMs will contribute to the natural evolution of language without causing collapse. This process could even be partially achieved in the near-term through prompting and curriculum learning \citep{jiang2024instruction}, the inclusion of richer multi-modal context windows, and frameworks such as algorithmic reparation \citep{davis2021algorithmic,wyllie2024fairness}. Once again, this will require more data and deliberate use by different communities to fight model collapse. Moreover, the worse the collapse, the more difficult it will be to reintroduce sufficient language to stabilise the data distribution. If any communities have a smaller margin for error and should be seeking such interventions, it is the low-resource and marginalized communities. 

\looseness=-1
\citet{schaeffer2025position} presents one important prior position paper which aimed to provide some needed precision in the discussion of model collapse. The primary point of \citet{schaeffer2025position} is that the literature on model collapse both conflates multiple definitions and over-simplifies the phenomenon. One of the primary critiques is that within a given paper, the definition of model collapse changes and distinct phenomena are used to characterize its occurrence. We fall into the same trap here: Sec.\ref{Sec:Environment} relies on a notion of model collapse that causes a change in neural scaling laws \citep[][and Definition 5 in \citet{schaeffer2025position}]{dohmatob2024tale}, while Sec.\ref{Sec:Culture} focuses mainly on the loss of data at the tails of the original distribution \citep[][and Definition 7 in \citet{schaeffer2025position}]{shumailov2023curse,wyllie2024fairness,shumailov2024ai}. It is not our aim here to argue for a particular definition, but rather to show that regardless of the definition used it is plausible that model collapse will disproportionately affect low-resource communities.

\looseness=-1
From the three ``types'' of model collapse definitions identified by \citet{schaeffer2025position}, we have mainly considered the consequences of two types. Specifically, we show that type 2, which considers the deformation of the data distribution, will likely first erase marginalized community data and promote the hegemonic viewpoints more so than even prior works considered \citep{bender2021dangers}. Secondly, we considered type 3, which is the decreasing value from additional data, and reconsidered the already alarming trend of marginalized communities disproportionately suffering the costs of environmental damage while benefiting the least from the new technologies \citep{strubell2019energy}. We extend this with the recent point from \citet{kandpalposition} claiming that data acquisition is the most expensive part of a training pipeline, and highlight that strategies revolving around collecting more data to mitigate model collapse will only be feasible for communities able to ``pay'' this data cost.

The type 1 definition of \citet{schaeffer2025position} is the progressive inability of the model to fit the original data distribution due to the introduction of synthetic data over generations. This type of definition is certainly important, particularly when lending theoretical understanding. However, this effect will also hit low-resource communities first as these data distributions will be more easily overwhelmed by synthetic data. Secondly, this type of definition provides a slightly more long-term perspective, while we are arguing for the immediate consideration of model collapse by low-resource communities. It is also worth noting that the intermediate generations are where the cultural harms noted in Sec.\ref{Sec:Culture} will arise, long before the generated content becomes incomprehensible and while it is still capable enough to influence public perception. Long before the models are unable to fit the original data distribution, we would argue that immense harm would have already been caused to low-resource communities through the disproportionate erasure of their data and the environmental damage caused by training sufficiently many models to reach this point. 

Thus, while \citet{schaeffer2025position} argue for a more discerning and muted rhetoric around model collapse, we are actually in agreement with their position, as they note that real tail data will be lost and that scaling laws may change with the introduction of synthetic data. They even note that ``loss of diversity is a real issue, with disproportionate harms often-times born by subgroups''---pre-empting our position in this work. We may even contextualize our position as a push towards a new axis of concern around model collapse: that model collapse will happen for different communities' data at different rates and have uneven impact. For example, \citet{schaeffer2025position} note that humanity may be able to train trillions of models before we notice the onset of model collapse. This conclusion was drawn based on a premise of there being a wealth of data available. This is certainly not the case for many communities and languages. Consequently, while the discourse around model collapse may be over-cautious and pessimistic, it is certainly accurate for some low-resource communities, regardless of which definition of model collapse is ultimately favoured. 

\looseness=-1
A final alternative view may consider whether the Stochastic Parrot analogy is still true, five years after the original work \citep{bender2021dangers}. While it is certainly true that immense progress has been made in recent years on grounded language understanding and generation \citep{radford2021learning,rombach2022high,alayrac2022flamingo}, particularly through the addition of multiple modalities, the statistical basis for the analogy remains true. LLMs base their predictions for the next token on statistical patterns which they have previously observed, which makes them susceptible to frequency and sample biases just like any other cognitive agent \citep{geva2025llms}. The question then becomes: to what extend does the additional modalities allow multimodal models to ground their language generations in reality rather than the statistics of the language they encounter? This is beyond the scope of this discussion, however it appears to be of paramount importance for future work. For now, we retain the Stochastic Parrot analogy as sufficiently relevant to provide the conceptual foundation for this work, particularly for low-resource languages and multilingual models, where clear sample biases and dominant patterns in the language data may not reflect broader societal realities \citep{bender2021dangers}. 



\section{Conclusions and Call to Action} \label{Sec:Call}
In this work we have aimed to contextualize the current discussions of model collapse \citep{shumailov2024ai,dohmatob2024model,dohmatob2024strong,bohacek2023nepotistically}
around prior discussions on the biases and injustices perpetuated by generative models and their training \citep{basta2019evaluating,bender2021dangers}. While many of the points raised here are not novel in isolation, it is the combination of perspectives we aim to contribute, as summarised in Tab.\ref{tab:summary-table}. While some works argue that model collapse may be less severe than the current literature would lead one to believe (and importantly caution against excessive rhetoric) \citep{schaeffer2025position}, we would argue that model collapse has a higher likelihood of reaching some communities which are already marginalized. \textit{Our position is that the degree of concern and concern should be based on the distribution, properties and quantity of a community's data}. Thus, whether one believes that model collapse is an imminent threat to our online cultures and progress in machine learning, it appears that low-resource and marginalized communities will be affected by it first and most severely. Consequently these communities should have a more prominent role in the academic discourse on the topic. 

Yet we find it troubling that this has yet to occur, with most experiments remaining on theoretical model  \citep{shumailov2024ai}, highly controlled datasets \citep{wyllie2024fairness} or high-resource settings \citep{wei2024understanding,wang2025bias}. While these highly impactful works have set the foundation for subsequent studies on model collapse, it cannot be assumed that preserving cultural sensitivity and enhancing representation of smaller communities aligns with the goals of the broader community. It is the responsibility of low-resource and marginalised communities to be proactive in maintaining the representation of their data and cultures as far as possible. Yet, we also note that the majority of people will form part of a minority group in some aspect. While axes such as race, gender and ethnicity rightly receive significant focus there are a number of different factors influencing a person's identity within their community. The homogenising effect of model collapse will affect all of these axes. Consequently, even if it is left to members within under-represented groups to understand and mitigate model collapse, model collapse still stands to be of wide-reaching concern for a large number of people. Thus, we conclude with a call to action: that the low-resource NLP and ML communities should take the forefront in understanding and mitigating model collapse. To this end, we provide some suggestions for open questions and first steps.

The open questions we highlight revolve primarily around the differences many low-resource languages present compared to some dominant languages such as English. Characterising the consequences of these differences on model collapse appears to be of utmost importance. For example, model collapse may manifest very differently on languages which are agglutinative (such as Turkish, Korean and Swahili) \citep{durrant2013formulaicity}. Agglutinative languages tend to form new words by the structured composition of other morphemes. In contrast to English, which is an analytic language (with some fusional elements) \citep{li2018transition} and has a few highly reusable words within the vocabulary, agglutinative languages have a large vocabulary and a very long tailed distribution of word use. While LLMs typically use sophisticated tokenizers which are able to extract the morphemes from agglutination \citep{kudo2018sentencepiece}, this still highlights the potential for differences between languages.
Even the choice of tokenizers, which have been shown to materially impact model performance on low-resource languages \citep{rajab2022effect}, could play a key role in mitigating model collapse for these languages. 

Another example of a unique property of low-resource languages is their propensity for code-switching \citep{olaleye2025afrocs}, which occurs when words from the vocabulary of one language are used inter-changeable in another. Code-switching, in general, currently poses a technological problem for low-resource languages \citep{winata2023decades,amol2025modelling} and invasive words will shift the data distribution. If invasive words push other words to the tails, then parts of the original language will be forgotten. Otherwise, the invasive words will occupy the tails and be removed. In both cases the redundancy acts to increase the tail of the distribution and will impact how model collapse occurs.

\looseness=-1
These examples highlight an important point: there are many reasons why a community's data may occupy the tails of the data distribution and be at risk of model collapse. While quantity of data is a clear contributor and saliently connected to marginalisation of communities, historical discrepancies in how technology is developed can also influence the data distribution even if a huge amount of data is acquired. The ability to influence the early design of foundational technology is also highly correlated with resource access and marginalisation. Since the underlying technologies behave differently between cultures and demographics, the established shift in data distribution from model collapse \citep{shumailov2024ai,wyllie2024fairness} will manifest as different tangible impacts on various cultures in the real world.

Turning now towards our recommendations, we can group these into three categories: 1) data, 2) detection, 3) decreasing. Regarding data, we agree with prior work on AI fairness that data curation remains an extremely valuable tool for avoiding bias and harm in LLMs \citep{rajab2025esethu}. The formation of datasets which have been sequestered and have guaranteed real data appears to be an imperative step  \citep{schaeffer2025position}. As \citet{schaeffer2025position} note, many of these datasets already exist from the work conducted by frontier AI labs and were created naturally through the development of the early LLMs and foundation models. This remains a solution only for the languages which have already seen significant development and are likely high-resource languages. Clearly, the existing range of well-curated datasets for thousands of low-resource languages \citep{agic2019jw300,siminyu2020ai4d,rajab2025esethu} would be insufficient to preserve the authenticity of the cultures they represent if model collapse were to become a larger issue. Thus, it should be the goal of policymakers, governments and academic institutions to construct such datasets which are guaranteed to contain real data and are representative of culture. How to systematically approach such a task remains a key open question.

The second category revolves around the clear need to both detect synthetic data within datasets, so that curation can be performed more accurately, and to be able to detect model collapse when it happens. In this case, we echo the important point raised by \citet{schaeffer2025position} that clear definitions are needed and that ideally support tractable measurements for model collapse or correlates of it. How feasible it is to detect synthetic data remains to be seen, and given the challenges already shown \citep{doughman2025exploring}, this should not be relied upon in isolation.

Finally, strategies to decrease model collapse would aim to fix the distributional shift which results from training on synthetic data. A number of strategies may be feasible and would follow from the first two steps once we understand the problem and have data which is usable. Some places to start may be in self-supervised learning or reinforcement learning to promote robust features \citep{hendrycks2019using} and out-of-distribution generalisation \citep{feng2024beyond}.

In conclusion, this work adds to the growing discussion and line of position papers \citep{farnadi2024position,kandpalposition,schaeffer2025position} related to model collapse. While uncertainty remains around the severity and prevalence of model collapse, we argue that this will differ between communities and disproportionately affect low-resource and marginalised communities. Thus, at least some of the most vulnerable communities will face the worst case scenario of model collapse. Consequently, our position is that significantly more effort from the low-resource ML and AI fairness communities is \textit{needed immediately} to understand how to measure and combat model collapse.



\section*{Acknowledgments}
SSM was supported by the Wallenberg AI, Autonomous Systems, and Software Program (WASP) and by OpenAI's Alignment Team, awarded by the UK AI Security Institute, through the Alignment Project. B.R. is a Canadian Institute for Advanced Research Fellow. Computations were performed using High Performance Computing infrastructure provided by the Mathematical Sciences Support unit at the University of the Witwatersrand.

\bibliography{example_paper}
\bibliographystyle{icml2026}
\end{document}

%% file: summary-table.tex
\begin{table*}[t]
\centering
\small
\begin{tabular}{@{}p{0.16\textwidth}p{0.35\textwidth}p{0.31\textwidth}p{0.11\textwidth}@{}}
\toprule
\textbf{Cause / mechanism} & \textbf{Why it matters for low-resource communities} & \textbf{Likely effect} & \textbf{Concerns} \\
\midrule
Synthetic-data feedback loops \citep{whitney2024real}&
Small real-data distributions are more easily overwhelmed by generated content. &
Dominant patterns are reinforced across generations, while rare linguistic and cultural forms are pushed further into the tail. &
Representation loss \\
\addlinespace
Tail erasure under model collapse \citep{wyllie2024fairness} &
Low-resource languages, dialects, and marginalized viewpoints often already occupy low-probability regions of the data distribution. &
Loss of diversity, reduced expressivity, and weaker model performance on communities whose data is already scarce. &
Cultural \newline erosion \\
\addlinespace
{Changed scaling \newline behaviour with \newline synthetic data} \citep{shumailov2024ai} &
If more data yields less improvement, communities with limited access to real data and compute have fewer viable routes to improvement. &
Higher data and compute requirements, widening the gap between high- and low-resource settings. &
Unequal \newline access \\
\addlinespace
High cost of real data acquisition \citep{kandpalposition} &
Collecting, validating, and maintaining culturally representative data is expensive and often unpaid or under-resourced. &
Low-resource communities may be pushed toward synthetic supplementation, increasing the risk of poisoning their own 
data. &
Data inequity \\
\addlinespace
Automated curation and synthetic-data \newline detection \citep{boutadjine2025human} &
Synthetic detection and filtering systems may be unreliable, opaque, or biased against uncommon language use. &
Authentic but rare or reclaimed language may be removed, while synthetic or hegemonic content remains. &
Curation bias \\
\addlinespace
Value-lock and hegemonic viewpoints \citep{bender2021dangers} &
LLMs tend to reproduce frequent historical patterns rather than socially evolving or locally grounded meanings. &
Outdated or dominant viewpoints persist, while new terms, norms, and movements struggle to enter the model distribution. &
Bias \newline reintroduction \\
\addlinespace
Multilingual transfer and language drift \citep{lu2020countering} &
Low-resource languages often depend on shared multilingual representations learned from high-resource languages. &
Fine-tuning or repeated generation may erode low-resource capabilities faster than high-resource ones. &
Language drift \\
\addlinespace
Tokenizer and morphology mismatch \citep{rajab2022effect} &
Agglutinative languages, code-switching, and non-standard orthographies may create longer-tailed token distributions. &
Model collapse may manifest differently across languages, making universal mitigation strategies unreliable. &
Measurement gap \\
\bottomrule
\end{tabular}
\vspace{0.15cm}
\caption{
Mechanisms by which model collapse may disproportionately affect low-resource and marginalized communities.
}
\label{tab:summary-table}
\vspace{-0.5cm}
\end{table*}

%% file: example_paper.bib
@article{kosmyna2025your,
  title={Your brain on ChatGPT: Accumulation of cognitive debt when using an AI assistant for essay writing task},
  author={Kosmyna, Nataliya and Hauptmann, Eugene and Yuan, Ye Tong and Situ, Jessica and Liao, Xian-Hao and Beresnitzky, Ashly Vivian and Braunstein, Iris and Maes, Pattie},
  journal={arXiv preprint arXiv:2506.08872},
  volume={4},
  year={2025}
}

@article{stankovic2025comment,
  title={Comment on: Your Brain on ChatGPT: Accumulation of Cognitive Debt When Using an AI Assistant for Essay Writing Tasks},
  author={Stankovic, Milos and Hirche, Ella and Kollatzsch, Sarah and Doetsch, Julia Nadine},
  journal={arXiv preprint arXiv:2601.00856},
  year={2025}
}

@inproceedings{wei2024understanding,
  title={Understanding the impact of ai-generated content on social media: The pixiv case},
  author={Wei, Yiluo and Tyson, Gareth},
  booktitle={Proceedings of the 32nd ACM International Conference on Multimedia},
  pages={6813--6822},
  year={2024}
}

@article{spennemann2025delving,
  title={Delving into: the quantification of Ai-generated content on the internet (synthetic data)},
  author={Spennemann, Dirk HR},
  journal={arXiv preprint arXiv:2504.08755},
  year={2025}
}

@article{shumailov2024ai,
  title={AI models collapse when trained on recursively generated data},
  author={Shumailov, Ilia and Shumaylov, Zakhar and Zhao, Yiren and Papernot, Nicolas and Anderson, Ross and Gal, Yarin},
  journal={Nature},
  volume={631},
  number={8022},
  pages={755--759},
  year={2024},
  publisher={Nature Publishing Group UK London}
}

@article{dohmatob2024model,
  title={Model collapse demystified: The case of regression},
  author={Dohmatob, Elvis and Feng, Yunzhen and Kempe, Julia},
  journal={Advances in Neural Information Processing Systems},
  volume={37},
  pages={46979--47013},
  year={2024}
}

@inproceedings{dohmatob2024tale,
  title={A Tale of Tails: Model Collapse as a Change of Scaling Laws},
  author={Dohmatob, Elvis and Feng, Yunzhen and Yang, Pu and Charton, Francois and Kempe, Julia},
  booktitle={International Conference on Machine Learning},
  pages={11165--11197},
  year={2024},
  organization={PMLR}
}

@inproceedings{gerstgrasser2024model,
  title={Is Model Collapse Inevitable? Breaking the Curse of Recursion by Accumulating Real and Synthetic Data},
  author={Gerstgrasser, Matthias and Schaeffer, Rylan and Dey, Apratim and Rafailov, Rafael and Korbak, Tomasz and Sleight, Henry and Agrawal, Rajashree and Hughes, John and Pai, Dhruv Bhandarkar and Gromov, Andrey and others},
  booktitle={First Conference on Language Modeling},
  year={2024}
}

@inproceedings{seddik2024bad,
  title={How bad is training on synthetic data? A statistical analysis of language model collapse},
  author={Seddik, Mohamed El Amine and Chen, Suei-Wen and Hayou, Soufiane and Youssef, Pierre and DEBBAH, Merouane Abdelkader},
  booktitle={First Conference on Language Modeling},
  year={2024}
}

@inproceedings{bohacek2023nepotistically,
  title={Nepotistically trained generative image models collapse},
  author={Bohacek, Maty and Farid, Hany},
  booktitle={ICLR 2025 Workshop on Navigating and Addressing Data Problems for Foundation Models},
  year={2023}
}

@inproceedings{silva2021towards,
  title={Towards a comprehensive understanding and accurate evaluation of societal biases in pre-trained transformers},
  author={Silva, Andrew and Tambwekar, Pradyumna and Gombolay, Matthew},
  booktitle={Proceedings of the 2021 Conference of the North American Chapter of the Association for Computational Linguistics: Human Language Technologies},
  pages={2383--2389},
  year={2021}
}

@inproceedings{ahn2022knowledge,
  title={Why knowledge distillation amplifies gender bias and how to mitigate from the perspective of DistilBERT},
  author={Ahn, Jaimeen and Lee, Hwaran and Kim, Jinhwa and Oh, Alice},
  booktitle={Proceedings of the 4th Workshop on Gender Bias in Natural Language Processing (GeBNLP)},
  pages={266--272},
  year={2022}
}

@article{odonnell2025math,
  title={We did the math on AI's energy footprint. Here's the story you haven't heard},
  author={O'Donnell, James and Crownhart, Casey},
  journal={MIT Technology Review},
  year={2025},
  month={May},
  day={20},
  url={https://www.technologyreview.com/2025/05/20/1116327/ai-energy-usage-climate-footprint-big-tech/}
}

@article{electricity2024analysis,
  title={Analysis and Forecast to 2026},
  author={Electricity, IEA},
  journal={International Energy Agency: Paris, France},
  year={2024}
}

@inproceedings{kirsten2025impact,
  title={The impact of inference acceleration on bias of llms},
  author={Kirsten, Elisabeth and Habernal, Ivan and Nanda, Vedant and Zafar, Muhammad Bilal},
  booktitle={Proceedings of the 2025 Conference of the Nations of the Americas Chapter of the Association for Computational Linguistics: Human Language Technologies (Volume 1: Long Papers)},
  pages={1834--1853},
  year={2025}
}

@article{gallegos2024bias,
  title={Bias and fairness in large language models: A survey},
  author={Gallegos, Isabel O and Rossi, Ryan A and Barrow, Joe and Tanjim, Md Mehrab and Kim, Sungchul and Dernoncourt, Franck and Yu, Tong and Zhang, Ruiyi and Ahmed, Nesreen K},
  journal={Computational Linguistics},
  volume={50},
  number={3},
  pages={1097--1179},
  year={2024},
  publisher={MIT Press 255 Main Street, 9th Floor, Cambridge, Massachusetts 02142, USA~…}
}

@inproceedings{rao2023ethical,
  title={Ethical reasoning over moral alignment: A case and framework for in-context ethical policies in LLMs},
  author={Rao, Abhinav Sukumar and Khandelwal, Aditi and Tanmay, Kumar and Agarwal, Utkarsh and Choudhury, Monojit},
  booktitle={Findings of the Association for Computational Linguistics: EMNLP 2023},
  pages={13370--13388},
  year={2023}
}

@inproceedings{farnadi2024position,
  author       = {Golnoosh Farnadi and
                  Mohammad Havaei and
                  Negar Rostamzadeh},
  title        = {Position: Cracking the Code of Cascading Disparity Towards Marginalized
                  Communities},
  booktitle    = {Forty-first International Conference on Machine Learning, {ICML} 2024,
                  Vienna, Austria, July 21-27, 2024},
  publisher    = {OpenReview.net},
  year         = {2024},
  url          = {https://openreview.net/forum?id=XDz9leJ9iK},
  bibsource    = {dblp computer science bibliography, https://dblp.org}
}

@inproceedings{bender2021dangers,
  title={On the dangers of stochastic parrots: Can language models be too big?},
  author={Bender, Emily M and Gebru, Timnit and McMillan-Major, Angelina and Shmitchell, Shmargaret},
  booktitle={Proceedings of the 2021 ACM conference on fairness, accountability, and transparency},
  pages={610--623},
  year={2021}
}

@inproceedings{samsi2023words,
  title={From words to watts: Benchmarking the energy costs of large language model inference},
  author={Samsi, Siddharth and Zhao, Dan and McDonald, Joseph and Li, Baolin and Michaleas, Adam and Jones, Michael and Bergeron, William and Kepner, Jeremy and Tiwari, Devesh and Gadepally, Vijay},
  booktitle={2023 IEEE High Performance Extreme Computing Conference (HPEC)},
  pages={1--9},
  year={2023},
  organization={IEEE}
}

@article{jegham2025hungry,
  title={How hungry is ai? benchmarking energy, water, and carbon footprint of llm inference},
  author={Jegham, Nidhal and Abdelatti, Marwan and Koh, Chan Young and Elmoubarki, Lassad and Hendawi, Abdeltawab},
  journal={arXiv preprint arXiv:2505.09598},
  year={2025}
}

@inproceedings{choi2024llm,
  title={The LLM effect: Are humans truly using LLMs, or are they being influenced by them instead?},
  author={Choi, Alexander and Akter, Syeda Sabrina and Singh, JP and Anastasopoulos, Antonios},
  booktitle={Proceedings of the 2024 Conference on Empirical Methods in Natural Language Processing},
  pages={22032--22054},
  year={2024}
}

@incollection{clark2004changing,
  title={Changing ideas about reference},
  author={Clark, Herbert H and Bangerter, Adrian},
  booktitle={Experimental pragmatics},
  pages={25--49},
  year={2004},
  publisher={Springer}
}

@article{clark2004speaking,
  title={Speaking while monitoring addressees for understanding},
  author={Clark, Herbert H and Krych, Meredyth A},
  journal={Journal of memory and language},
  volume={50},
  number={1},
  pages={62--81},
  year={2004},
  publisher={Elsevier}
}

@incollection{janzen2008intersubjectivity,
  title={Intersubjectivity in interpreted interactions: The interpreter's role in co-constructing meaning},
  author={Janzen, Terry and Shaffer, Barbara},
  booktitle={The shared mind: Perspectives on intersubjectivity},
  pages={333--355},
  year={2008},
  publisher={John Benjamins Publishing Company}
}

@article{chen2020artificial,
  title={Artificial intelligence in education: A review},
  author={Chen, Lijia and Chen, Pingping and Lin, Zhijian},
  journal={IEEE access},
  volume={8},
  pages={75264--75278},
  year={2020},
  publisher={Ieee}
}

@article{brynjolfsson2025generative,
  title={Generative AI at work},
  author={Brynjolfsson, Erik and Li, Danielle and Raymond, Lindsey},
  journal={The Quarterly Journal of Economics},
  volume={140},
  number={2},
  pages={889--942},
  year={2025},
  publisher={Oxford University Press}
}

@article{dohmatob2024strong,
  title={Strong model collapse},
  author={Dohmatob, Elvis and Feng, Yunzhen and Subramonian, Arjun and Kempe, Julia},
  journal={arXiv preprint arXiv:2410.04840},
  year={2024}
}

@article{shumailov2023curse,
  title={The Curse of Recursion: Training on Generated Data Makes Models Forget},
  author={Shumailov, Ilia and Shumaylov, Zakhar and Zhao, Yiren and Gal, Yarin and Papernot, Nicolas and Anderson, Ross J},
  journal={CoRR},
  year={2023}
}

@article{kaplan2020scaling,
  title={Scaling laws for neural language models},
  author={Kaplan, Jared and McCandlish, Sam and Henighan, Tom and Brown, Tom B and Chess, Benjamin and Child, Rewon and Gray, Scott and Radford, Alec and Wu, Jeffrey and Amodei, Dario},
  journal={arXiv preprint arXiv:2001.08361},
  year={2020}
}

@article{bahri2024explaining,
  title={Explaining neural scaling laws},
  author={Bahri, Yasaman and Dyer, Ethan and Kaplan, Jared and Lee, Jaehoon and Sharma, Utkarsh},
  journal={Proceedings of the National Academy of Sciences},
  volume={121},
  number={27},
  pages={e2311878121},
  year={2024},
  publisher={National Academy of Sciences}
}

@article{santos2022avoiding,
  title={Avoiding overfitting: A survey on regularization methods for convolutional neural networks},
  author={Santos, Claudio Filipi Gon{\c{c}}alves Dos and Papa, Jo{\~a}o Paulo},
  journal={ACM Computing Surveys (Csur)},
  volume={54},
  number={10s},
  pages={1--25},
  year={2022},
  publisher={ACM New York, NY}
}

@article{cogswell2015reducing,
  title={Reducing overfitting in deep networks by decorrelating representations},
  author={Cogswell, Michael and Ahmed, Faruk and Girshick, Ross and Zitnick, Larry and Batra, Dhruv},
  journal={arXiv preprint arXiv:1511.06068},
  year={2015}
}

@article{smith2003iterated,
  title={Iterated learning: A framework for the emergence of language},
  author={Smith, Kenny and Kirby, Simon and Brighton, Henry},
  journal={Artificial life},
  volume={9},
  number={4},
  pages={371--386},
  year={2003},
  publisher={MIT Press}
}

@article{kirby2008cumulative,
  title={Cumulative cultural evolution in the laboratory: An experimental approach to the origins of structure in human language},
  author={Kirby, Simon and Cornish, Hannah and Smith, Kenny},
  journal={Proceedings of the National Academy of Sciences},
  volume={105},
  number={31},
  pages={10681--10686},
  year={2008},
  publisher={National Academy of Sciences}
}

@article{kirby2014iterated,
  title={Iterated learning and the evolution of language},
  author={Kirby, Simon and Griffiths, Tom and Smith, Kenny},
  journal={Current opinion in neurobiology},
  volume={28},
  pages={108--114},
  year={2014},
  publisher={Elsevier}
}

@inproceedings{xia2024understanding,
  title={Understanding the performance and estimating the cost of llm fine-tuning},
  author={Xia, Yuchen and Kim, Jiho and Chen, Yuhan and Ye, Haojie and Kundu, Souvik and Hao, Cong Callie and Talati, Nishil},
  booktitle={2024 IEEE International Symposium on Workload Characterization (IISWC)},
  pages={210--223},
  year={2024},
  organization={IEEE}
}

@inproceedings{kandpalposition,
  title={Position: The Most Expensive Part of an LLM* should* be its Training Data},
  author={Kandpal, Nikhil and Raffel, Colin},
  year={2025},
  booktitle={Forty-second International Conference on Machine Learning Position Paper Track}
}

@inproceedings{strubell2019energy,
  title={Energy and policy considerations for deep learning in NLP},
  author={Strubell, Emma and Ganesh, Ananya and McCallum, Andrew},
  booktitle={Proceedings of the 57th annual meeting of the association for computational linguistics},
  pages={3645--3650},
  year={2019}
}

@article{vaswani2017attention,
  title={Attention is all you need},
  author={Vaswani, Ashish and Shazeer, Noam and Parmar, Niki and Uszkoreit, Jakob and Jones, Llion and Gomez, Aidan N and Kaiser, {\L}ukasz and Polosukhin, Illia},
  journal={Advances in neural information processing systems},
  volume={30},
  year={2017}
}

@article{xu2024survey,
  title={A Survey of Resource-efficient LLM and Multimodal Foundation Models},
  author={Xu, Mengwei and Yin, Wangsong and Cai, Dongqi and Yi, Rongjie and Xu, Daliang and Wang, Qipeng and Wu, Bingyang and Zhao, Yihao and Yang, Chen and Wang, Shihe and others},
  journal={CoRR},
  year={2024}
}

@inproceedings{devlin2019bert,
  title={Bert: Pre-training of deep bidirectional transformers for language understanding},
  author={Devlin, Jacob and Chang, Ming-Wei and Lee, Kenton and Toutanova, Kristina},
  booktitle={Proceedings of the 2019 conference of the North American chapter of the association for computational linguistics: human language technologies, volume 1 (long and short papers)},
  pages={4171--4186},
  year={2019}
}

@inproceedings{huangtowards,
  title={Towards Green AI in Fine-tuning Large Language Models via Adaptive Backpropagation},
  author={Huang, Kai and Yin, Hanyun and Huang, Heng and Gao, Wei},
  booktitle={The Twelfth International Conference on Learning Representations}
}

@article{iftikhar2025enhancing,
  title={Enhancing sustainability in LLM training: Leveraging federated learning and parameter-efficient fine-tuning},
  author={Iftikhar, Sunbal and Alsamhi, Saeed Hamood and Davy, Steven},
  journal={IEEE Transactions on Sustainable Computing},
  volume={10},
  number={6},
  pages={1158--1172},
  year={2025},
  publisher={IEEE}
}

@article{shi2025efficient,
  title={Efficient and Green Large Language Models for Software Engineering: Literature Review, Vision, and the Road Ahead},
  author={Shi, Jieke and Yang, Zhou and Lo, David},
  journal={ACM Transactions on Software Engineering and Methodology},
  volume={34},
  number={5},
  pages={1--22},
  year={2025},
  publisher={ACM New York, NY}
}

@article{ren2024reconciling,
  title={Reconciling the contrasting narratives on the environmental impact of large language models},
  author={Ren, Shaolei and Tomlinson, Bill and Black, Rebecca W and Torrance, Andrew W},
  journal={Scientific Reports},
  volume={14},
  number={1},
  pages={26310},
  year={2024},
  publisher={Nature Publishing Group UK London}
}

@inproceedings{basta2019evaluating,
  title={Evaluating the underlying gender bias in contextualized word embeddings},
  author={Basta, Christine and Costa-Juss{\`a}, Marta R and Casas, Noe},
  booktitle={Proceedings of the first workshop on gender bias in natural language processing},
  pages={33--39},
  year={2019}
}

@inproceedings{birhane2021large,
  title={Large image datasets: A pyrrhic win for computer vision?},
  author={Birhane, Abeba and Prabhu, Vinay Uday},
  booktitle={2021 IEEE Winter Conference on Applications of Computer Vision (WACV)},
  pages={1536--1546},
  year={2021},
  organization={IEEE}
}

@inproceedings{de2019does,
  title={Does object recognition work for everyone?},
  author={De Vries, Terrance and Misra, Ishan and Wang, Changhan and Van der Maaten, Laurens},
  booktitle={Proceedings of the IEEE/CVF conference on computer vision and pattern recognition workshops},
  pages={52--59},
  year={2019}
}

@inproceedings{kurita2019measuring,
  title={Measuring bias in contextualized word representations},
  author={Kurita, Keita and Vyas, Nidhi and Pareek, Ayush and Black, Alan W and Tsvetkov, Yulia},
  booktitle={Proceedings of the first workshop on gender bias in natural language processing},
  pages={166--172},
  year={2019}
}

@article{bao2024decoding,
  title={Decoding Matters: Addressing Amplification Bias and Homogeneity Issue for LLM-based Recommendation},
  author={Bao, Keqin and Zhang, Jizhi and Zhang, Yang and Huo, Xinyue and Chen, Chong and Feng, Fuli},
  journal={CoRR},
  year={2024}
}

@inproceedings{shin2024ask,
  title={Ask LLMs Directly,“What shapes your bias?”: Measuring Social Bias in Large Language Models},
  author={Shin, Jisu and Song, Hoyun and Lee, Huije and Jeong, Soyeong and Park, Jong C},
  booktitle={Findings of the Association for Computational Linguistics ACL 2024},
  pages={16122--16143},
  year={2024}
}

@inproceedings{jo2020lessons,
  title={Lessons from archives: Strategies for collecting sociocultural data in machine learning},
  author={Jo, Eun Seo and Gebru, Timnit},
  booktitle={Proceedings of the 2020 conference on fairness, accountability, and transparency},
  pages={306--316},
  year={2020}
}

@article{elkhatat2023evaluating,
  title={Evaluating the efficacy of AI content detection tools in differentiating between human and AI-generated text},
  author={Elkhatat, Ahmed M and Elsaid, Khaled and Almeer, Saeed},
  journal={International Journal for Educational Integrity},
  volume={19},
  number={1},
  pages={1--16},
  year={2023},
  publisher={Springer}
}

@article{ghiuruau2024distinguishing,
  title={Distinguishing reality from AI: approaches for detecting synthetic content},
  author={Ghiur{\u{a}}u, David and Popescu, Daniela Elena},
  journal={Computers},
  volume={14},
  number={1},
  pages={1},
  year={2024},
  publisher={MDPI}
}

@article{chaka2023detecting,
  title={Detecting AI content in responses generated by ChatGPT, YouChat, and Chatsonic: The case of five AI content detection tools},
  author={Chaka, Chaka},
  journal={Journal of Applied Learning and Teaching},
  volume={6},
  number={2},
  pages={94--104},
  year={2023}
}

@article{boutadjine2025human,
  title={Human vs. machine: A comparative study on the detection of AI-generated content},
  author={Boutadjine, Amal and Harrag, Fouzi and Shaalan, Khaled},
  journal={ACM Transactions on Asian and Low-Resource Language Information Processing},
  volume={24},
  number={2},
  pages={1--26},
  year={2025},
  publisher={ACM New York, NY}
}

@article{raffel2020exploring,
  title={Exploring the limits of transfer learning with a unified text-to-text transformer},
  author={Raffel, Colin and Shazeer, Noam and Roberts, Adam and Lee, Katherine and Narang, Sharan and Matena, Michael and Zhou, Yanqi and Li, Wei and Liu, Peter J},
  journal={Journal of machine learning research},
  volume={21},
  number={140},
  pages={1--67},
  year={2020}
}

@inproceedings{twyman2017black,
  title={Black Lives Matter in Wikipedia: Collective memory and collaboration around online social movements},
  author={Twyman, Marlon and Keegan, Brian C and Shaw, Aaron},
  booktitle={Proceedings of the 2017 acm conference on computer supported cooperative work and social computing},
  pages={1400--1412},
  year={2017}
}

@inproceedings{raj2024breaking,
  title={Breaking bias, building bridges: Evaluation and mitigation of social biases in llms via contact hypothesis},
  author={Raj, Chahat and Mukherjee, Anjishnu and Caliskan, Aylin and Anastasopoulos, Antonios and Zhu, Ziwei},
  booktitle={Proceedings of the AAAI/ACM Conference on AI, Ethics, and Society},
  volume={7},
  pages={1180--1189},
  year={2024}
}

@inproceedings{jiang2024instruction,
  title={Instruction-tuned language models are better knowledge learners},
  author={Jiang, Zhengbao and Sun, Zhiqing and Shi, Weijia and Rodriguez, Pedro and Zhou, Chunting and Neubig, Graham and Lin, Xi and Yih, Wen-tau and Iyer, Srini},
  booktitle={Proceedings of the 62nd Annual Meeting of the Association for Computational Linguistics (Volume 1: Long Papers)},
  pages={5421--5434},
  year={2024}
}

@inproceedings{lai2023chatgpt,
  title={Chatgpt beyond english: Towards a comprehensive evaluation of large language models in multilingual learning},
  author={Lai, Viet Dac and Ngo, Nghia and Veyseh, Amir Pouran Ben and Man, Hiếu and Dernoncourt, Franck and Bui, Trung and Nguyen, Thien Huu},
  booktitle={Findings of the association for computational linguistics: EMNLP 2023},
  pages={13171--13189},
  year={2023}
}

@inproceedings{lu2020countering,
  title={Countering language drift with seeded iterated learning},
  author={Lu, Yuchen and Singhal, Soumye and Strub, Florian and Courville, Aaron and Pietquin, Olivier},
  booktitle={International Conference on Machine Learning},
  pages={6437--6447},
  year={2020},
  organization={PMLR}
}

@inproceedings{lee2019countering,
  title={Countering Language Drift via Visual Grounding},
  author={Lee, Jason and Cho, Kyunghyun and Kiela, Douwe},
  booktitle={Proceedings of the 2019 Conference on Empirical Methods in Natural Language Processing and the 9th International Joint Conference on Natural Language Processing (EMNLP-IJCNLP)},
  pages={4385--4395},
  year={2019}
}

@inproceedings{hida2024social,
  title={Social Bias Evaluation for Large Language Models Requires Prompt Variations},
  author={Hida, Rem and Kaneko, Masahiro and Okazaki, Naoaki},
  booktitle={Findings of the Association for Computational Linguistics: EMNLP 2025},
  pages={14507--14530},
  year={2025}
}

@inproceedings{chisca2024prompting,
  title={Prompting fairness: Learning prompts for debiasing large language models},
  author={Chisca, Andrei-Victor and Rad, Andrei-Cristian and Lemnaru, Camelia},
  booktitle={Proceedings of the Fourth Workshop on Language Technology for Equality, Diversity, Inclusion},
  pages={52--62},
  year={2024}
}

@inproceedings{yang2023adept,
  title={Adept: A debiasing prompt framework},
  author={Yang, Ke and Yu, Charles and Fung, Yi R and Li, Manling and Ji, Heng},
  booktitle={Proceedings of the AAAI conference on artificial intelligence},
  volume={37},
  number={9},
  pages={10780--10788},
  year={2023}
}

@article{schaeffer2025position,
  title={Position: Model collapse does not mean what you think},
  author={Schaeffer, Rylan and Kazdan, Joshua and Arulandu, Alvan Caleb and Koyejo, Sanmi},
  journal={arXiv preprint arXiv:2503.03150},
  year={2025}
}

@inproceedings{wyllie2024fairness,
  title={Fairness feedback loops: training on synthetic data amplifies bias},
  author={Wyllie, Sierra and Shumailov, Ilia and Papernot, Nicolas},
  booktitle={Proceedings of the 2024 ACM Conference on Fairness, Accountability, and Transparency},
  pages={2113--2147},
  year={2024}
}

@article{davis2021algorithmic,
  title={Algorithmic reparation},
  author={Davis, Jenny L and Williams, Apryl and Yang, Michael W},
  journal={Big Data \& Society},
  volume={8},
  number={2},
  pages={20539517211044808},
  year={2021},
  publisher={SAGE Publications Sage UK: London, England}
}

@inproceedings{zhu2024synthesize,
  title={How to Synthesize Text Data without Model Collapse?},
  author={Zhu, Xuekai and Cheng, Daixuan and Li, Hengli and Zhang, Kaiyan and Hua, Ermo and Lv, Xingtai and Ding, Ning and Lin, Zhouhan and Zheng, Zilong and Zhou, Bowen},
  booktitle={Forty-second International Conference on Machine Learning},
  year={2025}
}

@inproceedings{alemohammad2023self,
  title={Self-consuming generative models go mad},
  author={Alemohammad, Sina and Casco-Rodriguez, Josue and Luzi, Lorenzo and Humayun, Ahmed Imtiaz and Babaei, Hossein and LeJeune, Daniel and Siahkoohi, Ali and Baraniuk, Richard},
  booktitle={The Twelfth International Conference on Learning Representations},
  year={2023}
}

@inproceedings{yoo2025code,
  title={Code-switching curriculum learning for multilingual transfer in llms},
  author={Yoo, Haneul and Park, Cheonbok and Yun, Sangdoo and Oh, Alice and Lee, Hwaran},
  booktitle={Findings of the Association for Computational Linguistics: ACL 2025},
  pages={7816--7836},
  year={2025}
}

@inproceedings{yoo2025code2,
  title={Code-switching red-teaming: Llm evaluation for safety and multilingual understanding},
  author={Yoo, Haneul and Yang, Yongjin and Lee, Hwaran},
  booktitle={Proceedings of the 63rd Annual Meeting of the Association for Computational Linguistics (Volume 1: Long Papers)},
  pages={13392--13413},
  year={2025}
}

@article{mohamed2025lost,
  title={Lost in the Mix: Evaluating LLM Understanding of Code-Switched Text},
  author={Mohamed, Amr and Zhang, Yang and Vazirgiannis, Michalis and Shang, Guokan},
  journal={arXiv preprint arXiv:2506.14012},
  year={2025}
}

@article{woolford1983bilingual,
  title={Bilingual code-switching and syntactic theory},
  author={Woolford, Ellen},
  journal={Linguistic inquiry},
  volume={14},
  number={3},
  pages={520--536},
  year={1983},
  publisher={JSTOR}
}

@inproceedings{kazdan2024collapse,
  title={Collapse or Thrive: Perils and Promises of Synthetic Data in a Self-Generating World},
  author={Kazdan, Joshua and Schaeffer, Rylan and Dey, Apratim and Gerstgrasser, Matthias and Rafailov, Rafael and Donoho, David L and Koyejo, Sanmi},
  booktitle={International Conference on Machine Learning},
  pages={29469--29494},
  year={2025},
  organization={PMLR}
}

@article{radford2019language,
  title={Language models are unsupervised multitask learners},
  author={Radford, Alec and Wu, Jeffrey and Child, Rewon and Luan, David and Amodei, Dario and Sutskever, Ilya and others},
  journal={OpenAI blog},
  volume={1},
  number={8},
  pages={9},
  year={2019}
}

@article{van2006impacts,
  title={The impacts of climate change on the risk of natural disasters},
  author={Van Aalst, Maarten K},
  journal={Disasters},
  volume={30},
  number={1},
  pages={5--18},
  year={2006},
  publisher={Wiley Online Library}
}

@inproceedings{bertrandstability,
  title={On the Stability of Iterative Retraining of Generative Models on their own Data},
  author={Bertrand, Quentin and Bose, Joey and Duplessis, Alexandre and Jiralerspong, Marco and Gidel, Gauthier},
  booktitle={The Twelfth International Conference on Learning Representations},
  year={2024}
}

@inproceedings{cui2025solvable,
  title={A solvable model of learning generative diffusion: theory and insights},
  author={Cui, Hugo and Pehlevan, Cengiz and Lu, Yue M},
  booktitle={The Thirty-ninth Annual Conference on Neural Information Processing Systems},
  year={2025}
}

@inproceedings{rajab2022effect,
  title={Effect of tokenisation strategies for low-resourced Southern African languages},
  author={Rajab, Jenalea},
  booktitle={3rd Workshop on African Natural Language Processing},
  year={2022}
}

@article{durrant2013formulaicity,
  title={Formulaicity in an agglutinating language: The case of Turkish},
  author={Durrant, Philip},
  year={2013},
  publisher={University of Exeter}
}

@article{li2018transition,
  title={The Transition from Comprehensive to Analytical Characteristics of English Language},
  author={Li, Xiaqing},
  journal={Theory and Practice in Language Studies},
  volume={8},
  number={9},
  pages={1241--1245},
  year={2018},
  publisher={Academy Publication Co., Ltd.}
}

@inproceedings{kudo2018sentencepiece,
  title={SentencePiece: A simple and language independent subword tokenizer and detokenizer for Neural Text Processing},
  author={Kudo, Taku and Richardson, John},
  booktitle={Proceedings of the 2018 Conference on Empirical Methods in Natural Language Processing: System Demonstrations},
  year={2018},
  organization={Association for Computational Linguistics}
}

@inproceedings{olaleye2025afrocs,
  title={AfroCS-xs: Creating a Compact, High-Quality, Human-Validated Code-Switched Dataset for African Languages},
  author={Olaleye, Kayode and Oncevay, Arturo and Sibue, Mathieu and Zondi, Nombuyiselo and Terblanche, Michelle and Mapikitla, Sibongile and Lastrucci, Richard and Smiley, Charese and Marivate, Vukosi},
  booktitle={Proceedings of the 63rd Annual Meeting of the Association for Computational Linguistics (Volume 1: Long Papers)},
  pages={33391--33410},
  year={2025}
}

@article{amol2025modelling,
  title={Modelling Misinformation in Swahili-English Code-switched Texts},
  author={Amol, Cynthia and Wanzare, Lilian and Obuhuma, James},
  journal={International Journal of Information Technology and Computer Science},
  volume={17},
  number={1},
  pages={67--80},
  year={2025}
}

@article{winata2023decades,
  title={The decades progress on code-switching research in NLP: A systematic survey on trends and challenges},
  author={Winata, Genta Indra and Aji, Alham Fikri and Yong, Zheng-Xin and Solorio, Thamar},
  journal={Findings of the Association for Computational Linguistics: ACL 2023},
  pages={2936--2978},
  year={2023}
}

@inproceedings{doughman2025exploring,
  title={Exploring the limitations of detecting machine-generated text},
  author={Doughman, Jad and Afzal, Osama Mohammed and Toyin, Hawau Olamide and Shehata, Shady and Nakov, Preslav and Talat, Zeerak},
  booktitle={Proceedings of the 31st International Conference on Computational Linguistics},
  pages={4274--4281},
  year={2025}
}

@article{geva2025llms,
  title={Do LLMs Exhibit Human-Like Cognitive Biases? A Large-Scale Systematic Evaluation},
  author={Geva, Tomer and Goldstein, Ariel and Lary, Eran and Levy, Coral},
  journal={A Large-Scale Systematic Evaluation (September 17, 2025)},
  year={2025}
}

@article{hendrycks2019using,
  title={Using self-supervised learning can improve model robustness and uncertainty},
  author={Hendrycks, Dan and Mazeika, Mantas and Kadavath, Saurav and Song, Dawn},
  journal={Advances in neural information processing systems},
  volume={32},
  year={2019}
}

@inproceedings{feng2024beyond,
  title={Beyond model collapse: Scaling up with synthesized data requires reinforcement},
  author={Feng, Yunzhen and Dohmatob, Elvis and Yang, Pu and Charton, Francois and Kempe, Julia},
  booktitle={ICML 2024 workshop on theoretical foundations of foundation models},
  year={2024}
}

@dataset{agentlans,
  author={agentlans},
  title={High Quality Multilingual Sentences},
  year={2025},
  publisher={Hugging Face},
  url={https://huggingface.co/datasets/agentlans/high-quality-multilingual-sentences}
}

@inproceedings{rombach2022high,
  title={High-resolution image synthesis with latent diffusion models},
  author={Rombach, Robin and Blattmann, Andreas and Lorenz, Dominik and Esser, Patrick and Ommer, Bj{\"o}rn},
  booktitle={Proceedings of the IEEE/CVF conference on computer vision and pattern recognition},
  pages={10684--10695},
  year={2022}
}

@article{zeighami2025cut,
  title={Cut Costs, Not Accuracy: LLM-Powered Data Processing with Guarantees},
  author={Zeighami, Sepanta and Shankar, Shreya and Parameswaran, Aditya},
  journal={Proceedings of the ACM on Management of Data},
  volume={3},
  number={6},
  pages={1--26},
  year={2025},
  publisher={ACM New York, NY, USA}
}

@article{jarvis2026compositionality,
  title={Compositionality and systematicity emerge from iterated learning in deep linear networks},
  author={Jarvis, Devon and Klein, Richard and Rosman, Benjamin and Saxe, Andrew M},
  journal={Proceedings of the National Academy of Sciences},
  volume={123},
  number={19},
  pages={e2509739123},
  year={2026},
  publisher={National Academy of Sciences}
}

@inproceedings{orgad2025llms,
  title={Llms know more than they show: On the intrinsic representation of llm hallucinations},
  author={Orgad, Hadas and Toker, Michael and Gekhman, Zorik and Reichart, Roi and Szpektor, Idan and Kotek, Hadas and Belinkov, Yonatan},
  booktitle={International Conference on Learning Representations},
  volume={2025},
  pages={66880--66913},
  year={2025}
}

@inproceedings{wilson2025bias,
  title={Bias Amplification in Stable Diffusion’s Representation of Stigma Through Skin Tones and Their Homogeneity},
  author={Wilson, Kyra and Ghosh, Sourojit and Caliskan, Aylin},
  booktitle={Proceedings of the AAAI/ACM Conference on AI, Ethics, and Society},
  volume={8},
  number={3},
  pages={2705--2717},
  year={2025}
}

@inproceedings{wang2025bias,
  title={Bias amplification: Large language models as increasingly biased media},
  author={Wang, Ze and Wu, Zekun and Zhang, Yichi and Guan, Xin and Jain, Navya and Lu, Qinyang and Gupta, Saloni and Koshiyama, Adriano},
  booktitle={Proceedings of the 14th International Joint Conference on Natural Language Processing and the 4th Conference of the Asia-Pacific Chapter of the Association for Computational Linguistics},
  pages={115--132},
  year={2025}
}

@article{alayrac2022flamingo,
  title={Flamingo: a visual language model for few-shot learning},
  author={Alayrac, Jean-Baptiste and Donahue, Jeff and Luc, Pauline and Miech, Antoine and Barr, Iain and Hasson, Yana and Lenc, Karel and Mensch, Arthur and Millican, Katherine and Reynolds, Malcolm and others},
  journal={Advances in neural information processing systems},
  volume={35},
  pages={23716--23736},
  year={2022}
}

@inproceedings{radford2021learning,
  title={Learning transferable visual models from natural language supervision},
  author={Radford, Alec and Kim, Jong Wook and Hallacy, Chris and Ramesh, Aditya and Goh, Gabriel and Agarwal, Sandhini and Sastry, Girish and Askell, Amanda and Mishkin, Pamela and Clark, Jack and others},
  booktitle={International conference on machine learning},
  pages={8748--8763},
  year={2021},
  organization={PmLR}
}

@inproceedings{agic2019jw300,
  title={JW300: A wide-coverage parallel corpus for low-resource languages},
  author={Agi{\'c}, {\v{Z}}eljko and Vuli{\'c}, Ivan},
  booktitle={Proceedings of the 57th Annual Meeting of the Association for Computational Linguistics},
  pages={3204--3210},
  year={2019}
}

@inproceedings{siminyu2020ai4d,
  title={AI4D-African Language Dataset Challenge},
  author={Siminyu, Kathleen and Freshia, Sackey},
  booktitle={Proceedings of the Fourth Widening Natural Language Processing Workshop},
  pages={68--77},
  year={2020}
}

@inproceedings{rajab2025esethu,
  title={The Esethu Framework: Reimagining Sustainable Dataset Governance and Curation for Low-Resource Languages},
  author={Rajab, Jenalea and Aremu, Anuoluwapo and Chimoto, Everlyn Asiko and Dunbar, Dale and Morrissey, Graham and Thior, Fadel and Potgieter, Luandrie and Ojo, Jessica and Tonja, Atnafu Lambebo and Nekoto, Wilhelmina NdapewaOnyothi and others},
  booktitle={Proceedings of the 63rd Annual Meeting of the Association for Computational Linguistics (Volume 1: Long Papers)},
  pages={30763--30776},
  year={2025}
}

@inproceedings{liu2025regmix,
  title={Regmix: Data mixture as regression for language model pre-training},
  author={Liu, Qian and Zheng, Xiaosen and Muennighoff, Niklas and Zeng, Guangtao and Dou, Longxu and Pang, Tianyu and Jiang, Jing and Lin, Min},
  booktitle={International Conference on Learning Representations},
  volume={2025},
  pages={38305--38339},
  year={2025}
}

@inproceedings{thudimixmin,
  title={MixMin: Finding Data Mixtures via Convex Minimization},
  author={Thudi, Anvith and Rovers, Evianne and Ruan, Yangjun and Thrush, Tristan and Maddison, Chris J},
  booktitle={Forty-second International Conference on Machine Learning}
}

@inproceedings{whitney2024real,
  title={Real risks of fake data: Synthetic data, diversity-washing and consent circumvention},
  author={Whitney, Cedric Deslandes and Norman, Justin},
  booktitle={Proceedings of the 2024 ACM conference on fairness, accountability, and transparency},
  pages={1733--1744},
  year={2024}
}
